\documentclass[journal]{IEEEtran}
\usepackage[table]{xcolor}
\usepackage{amsmath,amssymb,amsfonts}
\usepackage{graphicx}
\usepackage{subcaption}
\usepackage[hidelinks]{hyperref}
\usepackage{booktabs}
\usepackage{longtable}
\usepackage{tabularx}
\usepackage{ulem}
\usepackage{soul}

\renewcommand{\emph}[1]{\textit{#1}}

\newcommand{\sporc}{\texttt{SPoRC}}

\begin{document}

\title{\textit{Listening Between the Lines}: Decoding Podcast Narratives with Language Modeling}

\author{
  Shreya Gupta,
  Ojasva Saxena,
  Arghodeep Nandi,
  Sarah Masud,
  Kiran Garimella,
  and Tanmoy Chakraborty~\IEEEmembership{Senior Member, IEEE}%
  \thanks{S.G, O.S, A.N and T.C are with the Indian Institute Of Technology Delhi}
  \thanks{S.M is with the University of Copenhagen}
  \thanks{K.G is with Rutgers University}
  \thanks{Corresponding Authors: sarahmasud02@gmail.com}
}

\maketitle



\begin{abstract}

Podcasts have become a central arena for shaping public opinion, making them a vital source for understanding contemporary discourse. Their typically unscripted, multi-themed, and conversational style offers a rich but complex form of data. To analyze how podcasts persuade and inform, we must examine their narrative structures -- specifically, the narrative frames they employ.

The fluid and conversational nature of podcasts presents a significant challenge for automated analysis. We show that existing large language models, typically trained on more structured text such as news articles, struggle to capture the subtle cues that human listeners rely on to identify narrative frames. As a result, current approaches fall short of accurately analyzing podcast narratives at scale.

To solve this, we develop and evaluate a fine-tuned BERT model that explicitly links narrative frames to specific entities mentioned in the conversation, effectively grounding the abstract frame in concrete details. Our approach then uses these granular frame labels and correlates them with high-level topics to reveal broader discourse trends. The primary contributions of this paper are: (i) a novel frame-labeling methodology that more closely aligns with human judgment for messy, conversational data, and (ii) a new analysis that uncovers the systematic relationship between what is being discussed (the topic) and how it is being presented (the frame), offering a more robust framework for studying influence in digital media.
\end{abstract}


\maketitle

\section{Introduction}

\IEEEPARstart{P}{odcasts} have emerged as a powerful and decentralized platform for shaping public discourse, offering a unique medium for nuanced conversations and amplifying underrepresented voices \cite{aufderheide_podcasting_2020}. Their growing influence on political engagement is undeniable, yet this also brings significant concerns about their potential to spread misinformation and foster ideological polarization \cite{faculty_of_education_kampala_international_university_uganda_role_2025, cherumanal_everything_2024,10.1145/3746027.3754553}. The central problem is that while we recognize their impact, we lack effective methods to analyze how this influence is constructed at a narrative level. Understanding the persuasive strategies at play requires moving beyond what topics are discussed to how they are framed. This paper tackles the challenge of computationally identifying narrative frames within the vast and noisy landscape of podcasting to better understand the mechanisms of modern discourse.

This problem is complicated due to the inherent nature of podcast data. Unlike structured text such as news articles, podcast transcripts are conversational, subjective, and often lack clear rhetorical organization, making them challenging for downstream natural language processing tasks \cite{park_enhancing_2024, aquilina_end--end_2023}. Naive computational approaches, including the direct application of modern large language models (LLMs), often fail in this context. These models, while powerful, tend to prioritize statistically salient keywords over the subtle, contextual cues that human annotators use to identify narrative frames. For example, a zero-shot LLM might correctly identify a topic like ``climate change" but fail to distinguish whether it is being framed as an ``impending disaster," an ``economic opportunity," or a ``political conspiracy," a crucial distinction for understanding the speaker's intent and likely impact on the listener.

\begin{figure}[!t]
    \includegraphics[width=\columnwidth]{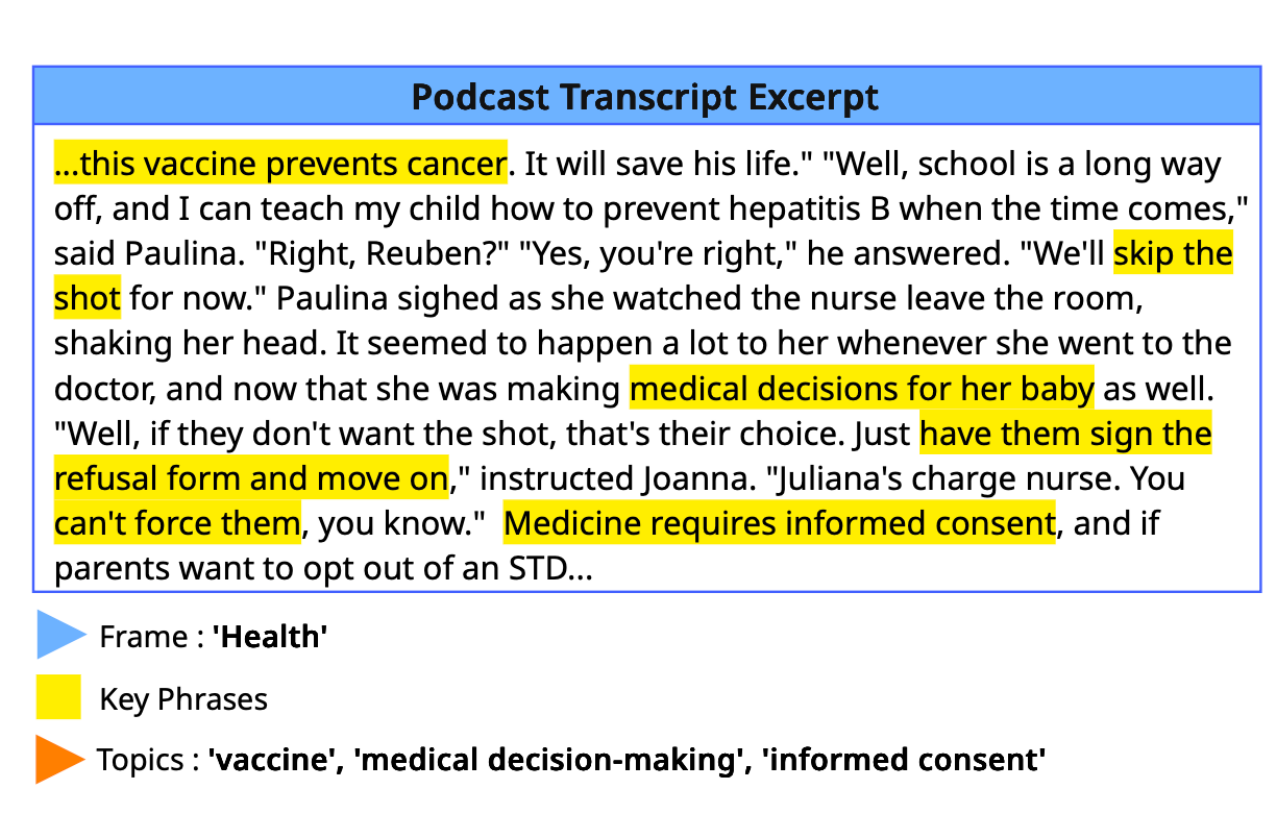}
    \caption{Sample \sporc\ transcript excerpt and its corresponding frame, key phrases, and topics.}
    \label{fig:transcript_example}
\end{figure}

Previous efforts to analyze podcasts at scale have produced valuable resources, such as the Structured Podcast Research Corpus (\sporc)~\cite{litterer_mapping_2025}. However, these often suffer from sparse or inconsistent annotations beyond basic transcriptions, limiting their utility for deep narrative analysis. Furthermore, most existing work applying LLMs to media analysis has not sufficiently addressed the unique challenges posed by conversational audio transcripts. Our work differs in that it overcomes the limitations of generic models. Instead of relying solely on statistical patterns in text, our primary innovation is to ground the abstract concept of a narrative frame in concrete, identifiable named entities. We hypothesize that linking a frame to the specific people, organizations, or locations provides the necessary context that generic models miss, leading to a more accurate and human-aligned analysis (Figure \ref{fig:transcript_example}).

To investigate this, we conduct a multi-faceted study on the \sporc\ transcript dataset. We first establish a baseline by evaluating a state-of-the-art LLM, Llama-3-8B-Instruct, on the task of zero-shot frame detection. We then develop and evaluate a multi-task, BERT-based model that is fine-tuned to jointly classify a narrative frame and its association with relevant named entities. Our approach systematically identifies dominant topics and entities (Section \ref{RQ1_method}), evaluates the strengths and critical failures of LLMs in frame detection (Section \ref{RQ2_method}), and demonstrates the superior performance of our fine-tuned, entity-aware model (Section \ref{RQ3_method}). 

Specifically, we make the following contributions: 

\begin{itemize}
\item We develop a nuanced approach for identifying influential named entities by leveraging temporal and topological features across podcast textual transcripts.
\item We present a technique for representing the interrelationships between dominant podcast topics, revealing clusters and thematic overlaps in discourse.
\item We evaluate Llama-3-8B Instruct for frame detection, identifying its strengths and limitations through comparison with manual annotations. Here, we highlight how LLMs fail to operationalize subjective aspects, unlike human annotators.
\item We contribute a multi-task setup fine-tuning an LLM for automatic frame detection and annotation assessment using frame labels and relevant named entity associations.
\end{itemize}

Our results confirm the limitations of general-purpose models, with Llama-3-8B achieving only 30-75\% accuracy across different frame types and exhibiting a clear bias towards salient features over contextual nuance. Critically, our experimental results indicate that fine-tuning a pretrained model with our entity-aware, multi-task approach yields a significant improvement of 5\%-15\% in classification accuracy across all frames. This demonstrates that while LLMs provide scalable solutions, a more targeted, fine-tuned approach is essential for accurately capturing the narrative structures that underpin the influential discourse taking place in today's podcast ecosystem.

\section{Related Work}
\textbf{Podcast Datasets.}
Early examples of conversational speech datasets for NLP tasks include the ATIS dataset \cite{hemphill_atis_1990} which focuses on air travel queries, and the NIST dataset \cite{garofolo_nist_2004}, which contains transcriptions of overlapping multi-speaker meetings. Despite featuring 700 unique speakers and 216 hours of transcripts, the TED-LIUM ASR corpus \cite{rousseau_ted-lium_2012} does not capture a multi-turn setup. Similarly, speech retrieval experiments using clean, continuous, single-speaker data have often employed the TIMIT corpora \cite{lopes_phoneme_2011}. However, the nature of audio-transcribed data poses challenges for exploring topics and themes in spoken content \cite{ishibashi_investigating_2020}. The setup is further complicated by the noisy, multi-speaker, multilingual long-form characteristics of podcasts \cite{jones_current_2021,federico_clef_2004}. To address this gap, the TREC Podcast Track \cite{jones_trec_2021} was launched with the release of the Spotify Podcast Dataset \cite{clifton_100000_2020}. TREC focused on two text-based tasks: retrieval of fixed two-minute segments and full-episode summarization. Recently, the Structured Podcast Research Corpus (\sporc) has been introduced \cite{litterer_mapping_2025}, containing transcriptions of 1.1M episodes from 247k shows. As \sporc\ is not curated for a single task, it enables a temporal and narrative-based analysis, such as ours.

\textbf{Topic Segmentation Methods.} Topic segmentation approaches fall broadly into two categories \cite{tur_topic_2011} -- (i) detecting explicit transitions, and (ii) identifying vocabulary shifts. The TextTiling algorithm \cite{hearst_text_1997} exemplifies the latter, detecting topical shifts through changes in vocabulary. Meanwhile, to capture the former, language model setups, such as BERTopic \cite{grootendorst_bertopic_2022} and TopicGPT \cite{reuter2024gptopicdynamicinteractivetopic}, have been recently proposed. BERTopic, applied to AI and healthcare, has outperformed traditional methods in podcast analysis \cite{grootendorst_bertopic_2022,dumbach_artificial_2024}. We employ the same for our analysis. Our work also highlights the extent of noise generated by BERTopic on large corpora, such as \sporc, and potential workarounds, as discussed in Section \ref{RQ1_method}. Readers are encouraged to reference the comprehensive benchmark of unsupervised topic segmentation models in \cite{gupta_comparative_2020}.

\textbf{Frame Detection Methods.} Framing can be defined as selecting and emphasizing elements of perceived reality (offline or online) to influence interpretation \cite{boydstun_tracking_nodate}.  Framing research primarily and largely relies on expert annotation \cite{eisele_capturing_2023,piskorski-etal-2025-semeval}. Such a setup is challenging to scale, especially for abstract social issues. To overcome this, unsupervised techniques, such as topic modeling, have been employed to detect transitions and trigger vocabulary, enabling the identification of frames \cite{walter_news_2019,tur_topic_2011}. Meanwhile, Bayesian and time series models have also been used for unsupervised frame detection \cite{tsur_frame_2015}.  

Computational advances have made large-scale frame analysis feasible via language modelling. This allows for a semi-supervised setup, where researchers trained binary classifiers per frame-topic pair using vocabulary indicators \cite{boydstun_tracking_nodate,jumle_finding_2024}. However, scalability continued to remain a challenge for multi-topic, multi-frame data. To further overcome annotation challenges, zero-shot prompting has been explored as a solution. Existing research on frame prediction for news articles has reported 43\% agreement between LLMs and human annotators  \cite{pastorino_decoding_2024,alonso_del_barrio_human_2024}. 

Our paper contributes to this discourse by investigating the potential of LLMs in detecting narrative frames in both large-scale and informal conversational texts, such as in \sporc. To the best of our knowledge, no prior work has systematically addressed the problem of frame detection in large-scale podcast corpora by explicitly modeling the relationship between narrative frames and the named entities they center on. The effectiveness of a fine-tuned, entity-aware model compared to zero-shot LLMs in this domain remains an open question, which this paper attempts to answer. Furthermore, while frame detection and topic modeling have been regularly employed to study narratives, they have been used as isolated techniques. Draw parallels in the real world; this study performs a temporal analysis of entities and frames, employing the two concepts in tandem to analyze emerging narratives.

\section{Dataset Preparation}
In this section, we profile \sporc\ \cite{litterer_mapping_2025} and describe the method for filtering the 1M dataset to a smaller, representative subset for computational feasibility. With a focus on textual transcripts, a list of feature subsets from \sporc\ relevant to the current study is given in Table \ref{tab:feature_descriptions}.

\begin{table}[htbp]
\centering
\renewcommand{\arraystretch}{1.2}
\scriptsize
\begin{tabular}{p{0.3\linewidth} p{0.63\linewidth}}
\toprule
\textbf{Feature} & \textbf{Description} \\
\midrule
\texttt{transcript} & Textual podcast data extracted from videos. \\
\texttt{podTitle} & Title of the podcast (not the same as episode title) \\
\texttt{epTitle} & Title of the particular episode for the podcast title \\
\texttt{oldestEpisodeDate} & Date of the oldest episode under that podcast title. \\
\texttt{Category labels} & Multi-label classification of the podcast topic, with up to 10 categories assigned to any podcast episode. \\
\texttt{durationSeconds} & Episode duration (in Seconds). \\
\texttt{episodeDate} & Date of episode release. \\
\bottomrule
\end{tabular}
\vspace{0.5em}
\captionsetup{labelfont=normalfont,textfont=normalfont}
\caption{A subset of columns of \sporc\ \cite{litterer_mapping_2025} used in the current study.}
\label{tab:feature_descriptions}
\end{table}

\begin{figure}[htbp]
    \includegraphics[width=0.5\textwidth, trim=0 0 0 0, clip]{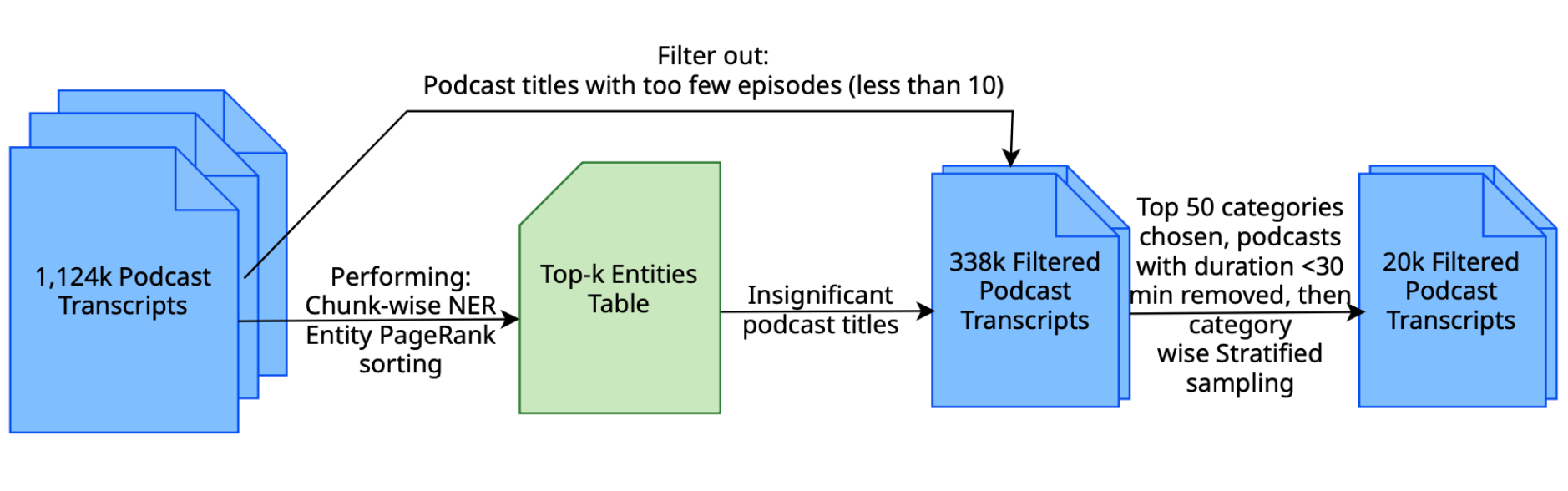}
    \caption{A multi-stage filtering approach to filter the podcasts.}
    \label{fig:data_prep}
\end{figure}

Figure \ref{fig:data_prep} illustrates our data preparation pipeline visually. We employ Named Entity Recognition (NER), temporal, and network features to obtain the most representative podcasts.

\subsubsection{Graph-based approach for selecting entities} 
To shortlist entities, we look beyond raw NER counts. Here, we construct an undirected, weighted bipartite network that connects podcasts to the entities they mention. The weight of the edge between a podcast and an entity is the number of times the podcast mentions the entity. The influential nodes are then identified based on the PageRank centrality \cite{zhang_pagerank_2021}. Entities are filtered to retain only the top 30,000 based on their node importance metrics (PageRank). A side-by-side comparison of top entities according to PageRank and simple entity counts can be seen in Table \ref{tab:pagerank_table}. Due to many entities in 1.1M samples, the PageRank values are in the order of $10^{-2}$ even for top-10 entities. The consistently high mentions of religious symbols occur because `religion' is the topmost podcast category. 

\subsubsection{Analysis of temporal distribution of podcasts}
On filtering the initial podcasts based on time between two consecutive podcasts ($>$ half a day) and minimum number of episodes in the title (\>=10), we obtain 338,407 podcasts. Further, the episodes are filtered based on minimum duration (141,493 podcasts), and only the top 50 categories are retained (140,471 podcasts). This is followed by sampling 60 episodes from titles with a large number of episodes (139,713 podcasts). Finally, from the category-wise stratified sampling of the 139,713 podcasts, a final set of 19,073 podcast episodes is obtained. \textit{From now on, when we mention \sporc\ or the data set, we refer to the 19k representative samples. }

\begin{table}[htbp]
\centering
\begin{tabular}{lccc} 
\toprule
\textbf{Entity} & \textbf{Pagerank} & \textbf{Entity} & \textbf{Count} \\
\midrule
Jesus       & 0.01507 & Jesus & 2660624 \\
Bible       &  0.004269 & Instagram &  608161 \\
Instagram   & 0.00374 & America & 559406 \\
COVID       &  0.00317 & Twitter & 510672 \\
America     & 0.003016 & COVID & 451510 \\
Youtube     & 0.002886 & John & 446512 \\
Twitter     & 0.002838 & Christian & 435348 \\
Christian   &  0.002532 & New York & 421028 \\
US          &  0.00283 & US & 414136 \\
\bottomrule
\end{tabular}
\vspace{0.5em}
\caption{A comparison of top-9 entities along with their PageRank scores and raw counts.}
\label{tab:pagerank_table}
\end{table}

\section{Methodology} 

In this section, we present our approach to analyzing the topical and framing characteristics of podcast discourse.

\begin{table*}[htbp]
\centering
\renewcommand{\arraystretch}{1.2}
\begin{tabular}{p{0.22\linewidth} p{0.70\linewidth}}
\toprule
\textbf{Textual Feature} & \textbf{Description} \\
\midrule
\textbf{Toxicity} & Toxicity measures the extent to which language is perceived as rude, hostile, or inflammatory. \\
\textbf{Sentiment} & Sentiment analysis evaluates the emotional valence of text—whether it is positive, negative, or neutral. \\
\textbf{Modality} & Modality refers to expressions of obligation, permission, or likelihood, often conveyed through modal verbs such as must, should, can, and may. \\

\textbf{Hedging} & Hedging involves language that softens assertions or introduces uncertainty, using terms like might, could, possibly, and somewhat. \\
\textbf{Degree Modifiers} & Degree modifiers (e.g., very, extremely, just) intensify or downplay the strength of statements. \\
\textbf{Part-of-Speech (PoS) Tags} & Analyzing PoS tags provides insight into the structural composition of texts. \\
\textbf{Frame-Based Vocabularies} & Each frame is associated with a distinctive set of keywords and phrases. \\
\bottomrule
\end{tabular}
\vspace{0.5em}
\caption{Description of significant extracted textual features from podcast transcript chunks.}
\label{tab:textual_features}
\end{table*}

\subsection{Entity-Topic Analysis} 
\label{RQ1_method}
The first question to investigate is ``what entities and topics are discussed in the podcasts.'' We start with zero-shot LLM prompting for topic detection. As its inference process is slow\footnote{It took $\approx$18 days to classify chunks of 5000 podcasts.}, we employ BERTopic  \cite{grootendorst_bertopic_2022} models instead. BERTopic runs on 250-token chunks of podcast text. We determine the chunk size through preliminary experiments with BERTopic conducted on a smaller subset of podcasts. Interestingly, when we apply the chunk-based topic model to the 19k podcasts, it yields incorrect topic associations. Hence, category-wise (based on category labels assigned in \sporc), 250 token chunk-based topic modeling is performed. The setup strikes a balance between contextual richness and computational efficiency, thereby optimizing the model's performance.

\subsection{Frame Prediction} \label{RQ2_method}
We now use framing to understand better the intent and narratives of how the topics and entities are being discussed. Frames constitute the perspectives or lenses through which content is presented \cite{van_hulst_discourse_2025}. Existing methods for frame detection rely on a vocabulary, or specifically trained frame-wise classifiers {\cite{boydstun_tracking_nodate}, or some subjective technique \cite{hutchison_automatic_2013}. However, none of these techniques is scalable and versatile enough for a large dataset like ours. We thus experiment with LLM prompting. Llama 8B Instruct \cite{noauthor_meta-llamameta-llama-3-8b-instruct_2024} is prompted on a subset of 6k chunks to assign one of six predefined frame categories to chunks of podcast transcripts: \textit{Social, Health, Legal, Financial, Security, and Moral}. Figure \ref{fig:LLM_frame_prompt} enlists the prompting setup.

However, due to the subjective nature of the task, even LLMs do not consistently achieve high accuracy across all frame types. We conduct an extensive error analysis to gain a deeper understanding of these limitations.  

\begin{figure}[!t]
    \centering
    \includegraphics[width=0.9\columnwidth]{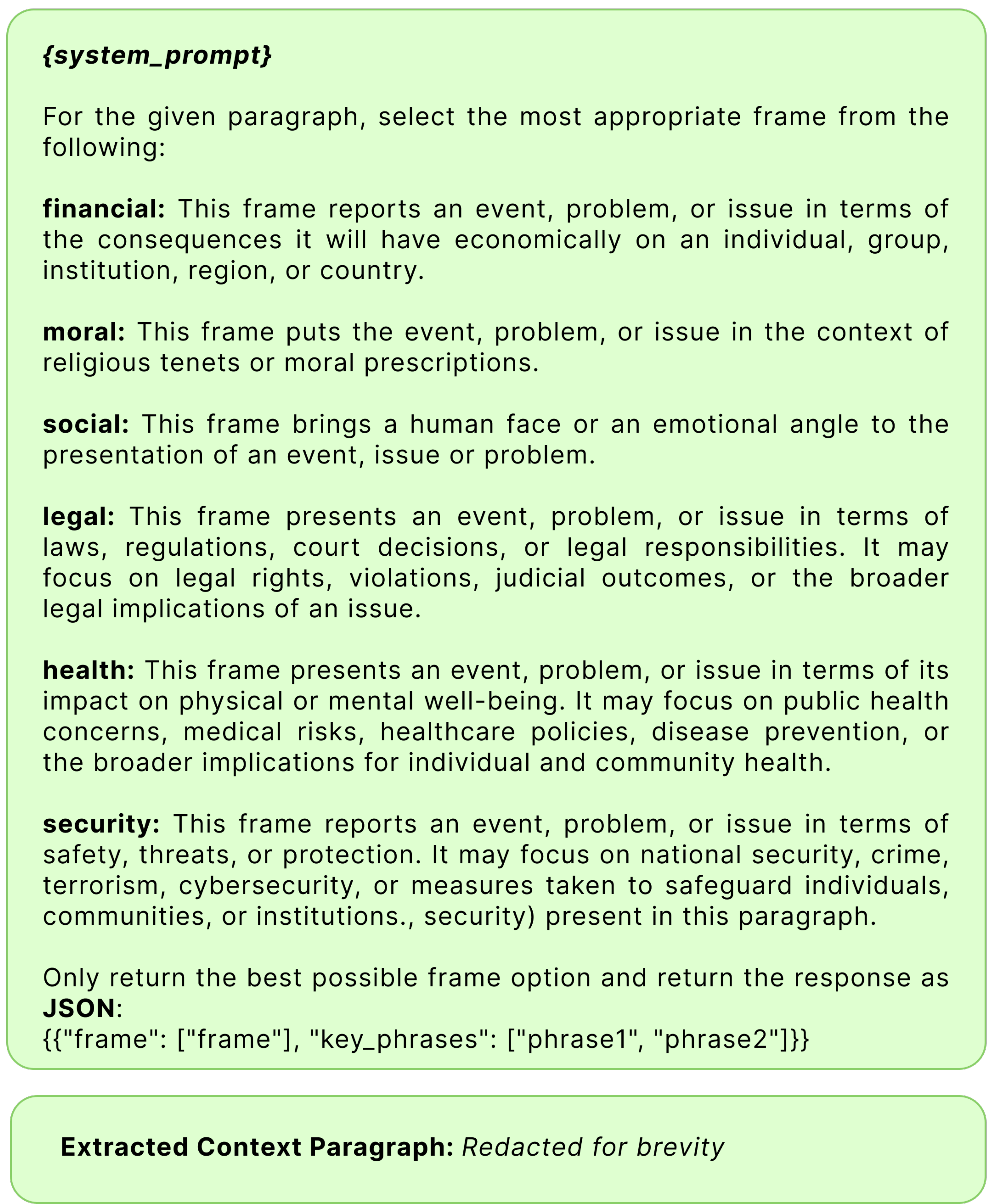}
    \caption{Prompt to extract frames.}
    \label{fig:LLM_frame_prompt}
\end{figure}

\subsection{Manual Evaluation} \label{RQ2_analysis}
To better understand the LLM's shortcomings, we manually review a random subset of 600 frame predictions (100 per frame type) by the LLM. An expert annotator\footnote{The expert is not affiliated with any political organization or endorses any podcast examined in this study.} from the field of CSS volunteered to review the frame labels and key phrases (rationales) generated by the LLM. The evaluator also documents corrections and the error patterns. This enables us to examine the difference in the importance of textual features between LLM-predicted and human-annotated classifications by training supervised models on both annotation sources.

\paragraph*{Textual feature extraction} Based on our observations, we extract a comprehensive set of textual features spanning abstract elements like toxicity and sentiment, and objective elements like part-of-speech tag counts. Table \ref{tab:textual_features} lists the textual features employed in this setup. 

\paragraph*{Textual feature importance}
The subset of incorrect LLM labels and their extracted features is used to train a random forest and a logistic regression model. Separate models are built for the human-annotated and LLM-annotated sets to understand the importance of the general feature. 

\subsection{Fine-Tuned Frame Prediction} \label{RQ3_method}
Owing to the performance gaps in LLM annotations and the low inference speed, we investigate whether a BERT-based PLM can yield better frame accuracies. Existing literature has reported the better task-specificity of fine-tuned PLMs over LLMs \cite{yadav-etal-2024-tox,upravitelev-etal-2025-comparing}, and we examine the same for frame detection. 

\paragraph*{Multi-task BERT}
Using a multi-task setup, we attempt to capture the influence of the \textit{`rationales'} (key phrases) that serve as indicators of specific frames. To this end, BERT is trained on two objectives: span detection for key phrases/rationales and frame classification. The model is evaluated over 30 epochs, and the confusion matrix for each epoch is analyzed to understand which frame types are more complex to distinguish between. This fine-tuned model is then employed to frame all the podcast transcript chunks. Standard label encoding is used for the frame classification task, while the span detection task utilizes B-I-O (Begin-Inside-Outside) labels to represent the location of key phrases. Cross-entropy loss is employed for training both tasks. The model is fine-tuned on the dataset comprising 600 human-annotated chunks.

\paragraph*{Assessing fine-tuned BERT model} \label{sec:bert_methodology}
As manual annotation for direct ground-truth evaluation of the 760k podcast chunks (spanning 19k podcasts) is infeasible, we design an indirect assessment methodology (see \ref{sec:bert_methodology}) based on the distribution of frames over particular named entities. The hypothesis guiding this assessment is that certain named entities (e.g., “COVID”, “Jesus Christ”) are contextually associated with specific frames (e.g., “Health”, “Morality”) due to real-world patterns. The important named entities (as described in Section \ref{RQ1_method}) are manually selected to associate relevant frames corresponding to each entity, thereby assessing the automatic frame classification. Regular expression matching is used for both exact matches and closely related variations of the entity to determine the presence of a named entity within a podcast segment.

\subsection{Models Used} \label{subsection:models}
Here, we describe the various models used in this study. 
\begin{itemize}
 \item \textbf{Spacy.} `Spacy en\_core\_web\_sm', is part of the Spacy library with an English pipeline optimized for CPU. It is employed for NER and PoS Tagging.
 \item \textbf{Llama 3.} The Llama 3-8B Instruct \cite{noauthor_meta-llamameta-llama-3-8b-instruct_2024} is optimized for dialogue use cases, and is employed here for prompting frame classification. We employ the HuggingFace version of the model.
    \item \textbf{Toxicity scores.} The `unbiased-toxic-roberta' model on HuggingFace has been trained on the Jigsaw toxicity dataset \cite{noauthor_unitaryunbiased-toxic-roberta_nodate}.  The model extracted toxicity scores, one of the many textual features.
    \item \textbf{VADER.} VADER (Valence Aware Dictionary and sEntiment Reasoner) is a lexicon and rule-based sentiment analysis tool attuned explicitly to sentiments expressed in social media. This model is used to extract sentiment-based features \cite{hutto_vader_2014}.
    \item \textbf{BERT.} The `bert-base-uncased' \cite{devlin_bert_2018} from HuggingFace  is used for frame classification.
\end{itemize}

\section{Observations \& Results}

This section presents the findings at each stage of the analysis. Our discussion encompasses the expected increase in entity counts resulting from real-world events, as well as secondary effects on the emergence of podcast discussions that are not directly related to these events. We conclude this section with a discussion of a three-way comparison of the zero-shot LLM, human annotator, and fine-tuned PLM efficacies, highlighting the scope of each. 

\subsection{Entity Count Analysis over Time}
We map the NER counts to the days episodes were released to understand how entity frequency changed over time. As the podcasts span May-June 2020, we observe a 10-fold increase in mentions of entities like COVID-19 as the virus spreads. During the same period, speculation about the connection between COVID-19 and China increased, as reflected in the growing mentions of China over time. 

\subsection{Topic-Word Correlations} 
\label{sec:topic-word} 
Beyond NER mentions, category-wise BERTopic \cite{grootendorst_bertopic_2022} helps draw insights into how podcasts interlace with socio-political events.

\paragraph{Socio-political categories}
An analysis of topics directly linked to the socio-political landscape, including daily news, commentary, and politics, reveals a high correlation with current affairs. It can be seen in the mention of George Floyd, Biden, Black voters, and geopolitical issues in Iran, Syria, and China (due to rising COVID cases). Unsurprisingly, one of the top-mentioned topics in daily news is the intersection of protests and police, along with the issue of wearing masks during the COVID-19 pandemic. We also observe the mention of China in relation to the Hong Kong-mainland China conflict that began in late 2019 and continued until early 2020. Political podcasts, on the other hand, have topics like defunding the police, Brexit, and healthcare, along with the mention of Antifa and a nuclear treaty. All of these show that the podcasts are, in fact, reflecting and commenting on current affairs. 

\paragraph{Racism in Zeitgeist}
The football topics also feature Colin Kaepernick kneeling as a top topic \cite{noauthor_timeline_nodate}. 
Interestingly, correlations of racism, `black', and `white' are observed in a broader range of categories, including religion, careers, video games, and sports that traditionally have a prominent Black representation. \textbf{This shows that even podcasts on topics like `beauty' and `fashion', which are generally not perceived as necessary for analyzing the socio-political landscape, still discuss socially charged topics.}

\paragraph{Impact of COVID-19 on daily life}
The impact of COVID-19 is evident in the consistent mention of concepts such as ``lockdown'' and ``COVID tests'' in basketball and football podcasts. Meanwhile, issues such as `remote working, masks, and vaccination' become visible in business and the comedy genre. `Virus, cases, and testing' are evident in government-related podcasts. Additionally, {\textbf{secondary effects such as distance learning are mentioned in family-themed podcasts. It indicates that the pandemic had a significant impact across different fields, either in a primary or a secondary manner.}}  

\begin{figure}[!t]
    \includegraphics[width=0.47\textwidth]{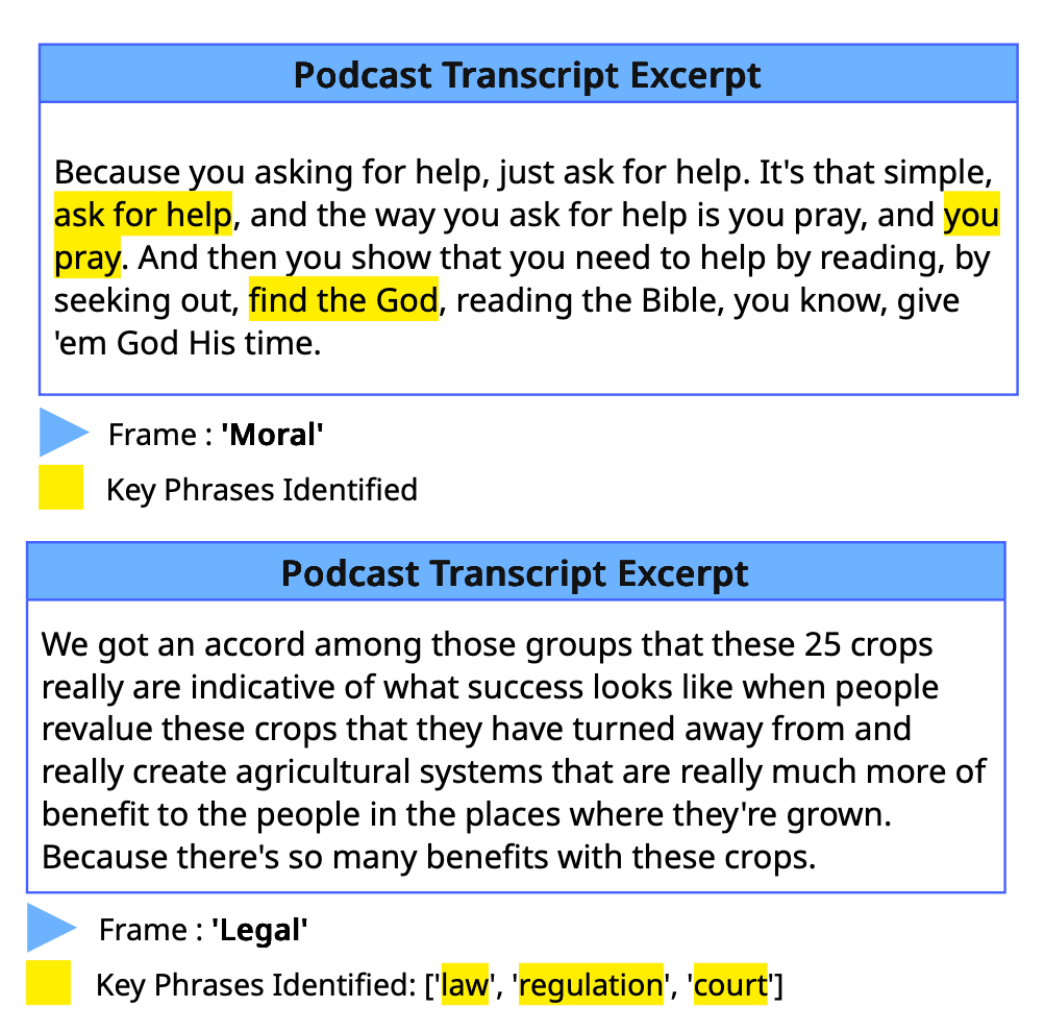}
    \caption{LLM annotations (bottom) for podcast chunks as compared to a correct LLM annotation (top). The LLM outputs hallucinated key phrases for the bottom chunk.}
    \label{fig:manual_annotation}
\end{figure}

\subsection{Framing Results and Limitations}
The NER analysis helps identify the various entities and how their mention evolves. BERTopic further helps correlate entities based on occurrence patterns. While topic modeling and NER analysis give a good insight into \textit{what} is discussed, they are inadequate in providing clues on \textit{how} things are said. Therefore, it was essential to further identify the frames employed for different concepts to analyze the intent behind the narratives in the podcasts. 

The LLM output is post-processed to determine the frame predicted for each chunk. After classification, the filtering ensures that at least 1000 chunks are present per frame. Looking closely at Figure \ref{fig:manual_annotation}, we realize that in some cases the key phrases identified by the LLM do not exist in the podcast chunk at all. Here, the LLM hallucinates key phrases to justify the misclassified frame label. Expert annotators also record noteworthy patterns during the annotation process, which we discuss in Section \ref{sec:discussion}. Based on the manual annotations of 600 frame chunks, the per-category accuracy for both frame labels and key phrases is reported in Table \ref{tab:epochs_table}. {\textbf{LLM accuracies vary across frames, with the `health' frame showing the lowest accuracy at 41\%, and the `social' frame achieving the highest accuracy at 76\%.}}

\begin{figure}[htbp]
  \centering
  \begin{subfigure}{\linewidth}
    \centering
    \includegraphics[width=0.95\linewidth]{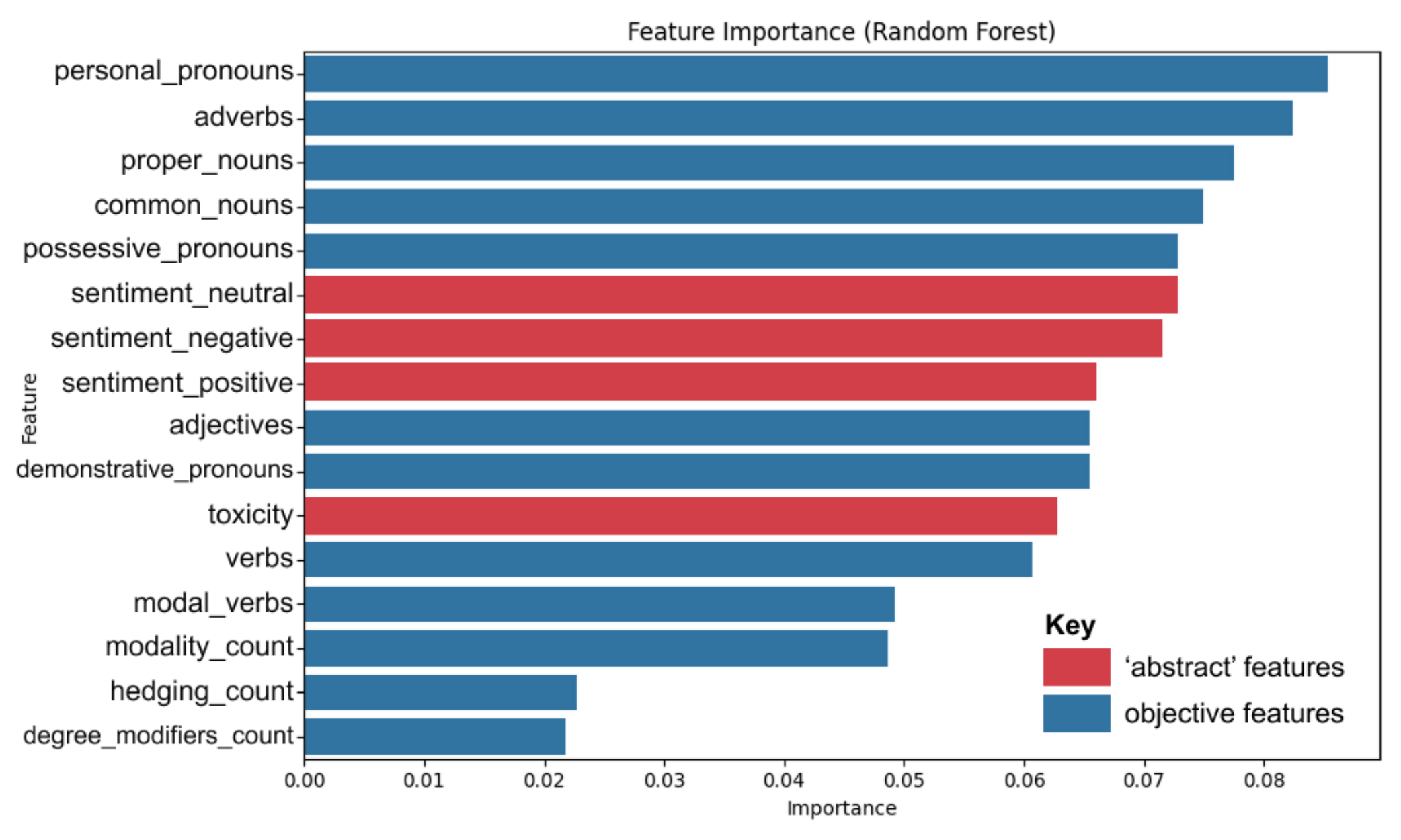}
    \caption{Textual feature priority: LLM annotations (RF classifier)}
  \end{subfigure}

  \vspace{1em}

  \begin{subfigure}{\linewidth}
    \centering
    \includegraphics[width=0.95\linewidth]{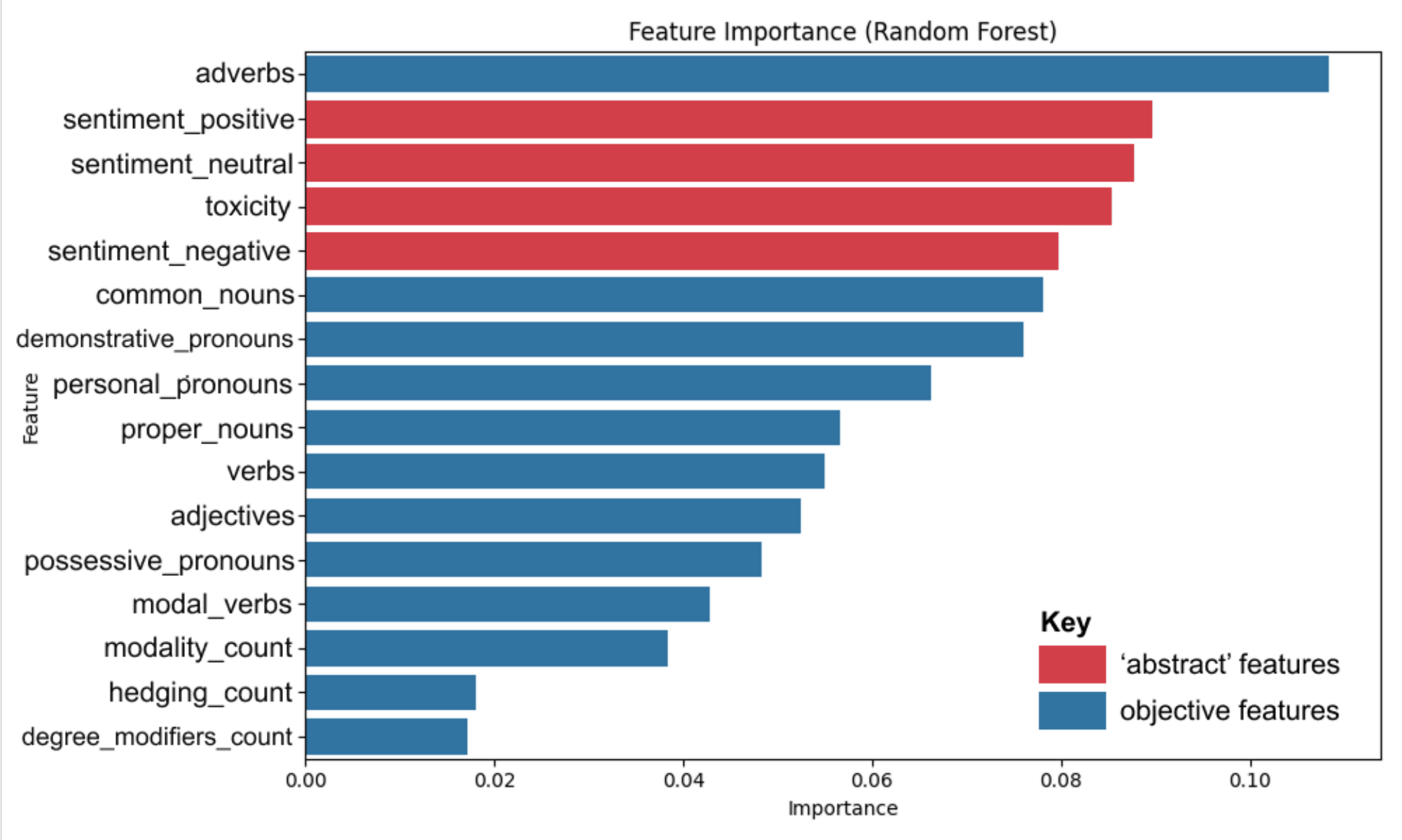}
    \caption{Textual feature priority: Human annotations (RF classifier)}
  \end{subfigure}
  \vspace{0.5em}
  \caption{Textual feature analysis shows that LLMs prioritize objective and statistical features (shown in blue). In contrast, human annotations emphasize more nuanced aspects such as toxicity and sentiment (shown in red).}
  \label{fig:feture_importance_all}
\end{figure}

\begin{table*}[ht]
\centering
\resizebox{\textwidth}{!}{%
\begin{tabular}{l|c|ccc|ccc|ccc|ccc|ccc}
\hline
\textbf{Frame} & \textbf{Accuracy} & \multicolumn{3}{c|}{\textbf{Epoch 5}} & \multicolumn{3}{c|}{\textbf{Epoch 10}} & \multicolumn{3}{c|}{\textbf{Epoch 15}} & \multicolumn{3}{c|}{\textbf{Epoch 20}} & \multicolumn{3}{c}{\textbf{Epoch 30}} \\
& & Prec & Acc & F1 & Prec & Acc & F1 & Prec & Acc & F1 & Prec & Acc & F1 & Prec & Acc & F1 \\
\hline
financial & 0.500 & 0.750 & 0.214 & 0.333 & 0.615 & \textbf{0.571} & 0.593 & 0.600 & 0.429 & 0.500 & 0.667 & 0.429 & 0.522 & 0.750 & 0.429 & 0.545 \\
health    & 0.410 & 0.750 & \textbf{0.750} & 0.750 & 0.857 & \textbf{0.750} & 0.800 & 0.857 & \textbf{0.750} & 0.750 & 0.857 & \textbf{0.750} & 0.800 & 0.538 & \textbf{0.875} & 0.667 \\
legal     & 0.320 & 1.000 & \textbf{0.667} & 0.800 & 1.000 & \textbf{0.778} & 0.875 & 0.875 & \textbf{0.778} & 0.778 & 0.875 & \textbf{0.778} & 0.824 & 0.727 & \textbf{0.889} & 0.800 \\
moral     & 0.720 & 0.846 & 0.579 & 0.688 & 0.882 & \textbf{0.789} & 0.833 & 0.867 & 0.684 & 0.764 & 0.765 & \textbf{0.882} & 0.833 & 1.000 & 0.316 & 0.480 \\
security  & 0.590 & 0.917 & 0.458 & 0.611 & 0.750 & 0.500 & 0.600 & 0.765 & 0.542 & 0.634 & 0.765 & 0.542 & 0.634 & 0.778 & 0.583 & 0.667 \\
social    & 0.760 & 0.560 & \textbf{0.955} & 0.706 & 0.672 & \textbf{0.886} & 0.765 & 0.639 & \textbf{0.886} & 0.743 & 0.667 & \textbf{0.909} & 0.769 & 0.597 & \textbf{0.841} & 0.698 \\
\hline
\end{tabular}
}
\vspace{0.5em}
\caption{Frame-wise evaluation of our BERT-style encoder, showing Precision, Accuracy, and F1 progression across training epochs (5, 10, 15, 20, and 30). Ground-truth label accuracy provides a baseline performance.}
\label{tab:epochs_table}
\end{table*}

\subsection{Human vs LLM Feature Importance}
A set of 270 texts incorrectly labeled by the LLM (compared to ground-truth labels) is used to train three models: Random Forest, Logistic Regression, and One-vs-Rest Classifiers. Although the classification accuracies were modest, ranging from 30\% to 60\%, notable differences are evident in the feature importance rankings between the two annotation sources. As shown in Figure \ref{fig:feture_importance_all}, human annotations tend to assign higher importance to abstract features (depicted in red) such as sentiment and toxicity. In contrast, LLM annotations emphasize more objective features (shown in blue), such as counts of specific PoS tags. \textbf{It indicates a more formulaic classification strategy in the case of LLMs.} 

We further use one-vs-rest analysis to obtain frame-specific feature importance rankings. We train six binary classifiers, each targeting the classification of one frame type against all others. This approach provides granular insights into the discriminative features associated with individual frames. Some interesting findings included the higher feature importance given to positive sentiment in the case of social frames, as well as the higher importance of personal pronouns (with coefficients between 0.5 and 0.6 for both human annotations). Similarly, for the security frames, there is a high positive correlation with negative sentiment (ranging from 0.6 to 0.7) and toxicity (ranging from 0.2 to 0.3).  

\begin{table*}[!t]
\centering
\begin{tabular}{p{5cm} p{3cm} p{6cm}}
\toprule
\textbf{Theme} & \textbf{Correlation} & \textbf{Speculations} \\
\midrule
\rowcolor{lightgray}
\multicolumn{3}{l}{\textbf{COVID in Health Domain}} \\
insurance\_healthcare & \textbf{0.633} & notable correlation with the healthcare and insurance response during the pandemic \\
trauma\_ptsd\_therapist & \textbf{0.625} & mental health issues like PTSD became more prominent during COVID \\
anxiety\_anxious\_weak & \textbf{0.624} & widespread anxiety and psychological vulnerability due to the pandemic \\
loneliness\_lonely\_franco & \textbf{0.617} & loneliness due to isolation and distancing measures \\
\rowcolor{lightgray}
\multicolumn{3}{l}{\textbf{Kaepernick in Football Realm}} \\
lack\_police\_people & \textbf{0.702} & police involvement and the people’s response to the Kaepernick controversy \\
quarterbacks\_brady\_quarte & 0.561 & higher correlation to the police as compared to the Karpernick's own position on the field \\
\rowcolor{lightgray}
\multicolumn{3}{l}{\textbf{Racism in Sports Realm}} \\
police\_cops\_black & \textbf{0.733} & underscores tensions between policing and Black athletes or fans within sports settings \\ coach\_basketball\_coaching & \textbf{0.703} & strong correlation with basketball and coaching suggests racialized understandings of talent and leadership \\
nba\_protests\_black & \textbf{0.788} & highlights the NBA’s central role in athlete-led protests against racial injustice \\
jordan\_michael\_lebron & 0.584 & surprisingly moderate -- iconic Black athletes are mentioned less in direct connection with racial activism than expected \\
\bottomrule
\end{tabular}
\vspace{0.5em}
\caption{Correlations between various entities across different themes.}
\label{tab:correlation_analysis_themes_short}
\end{table*}

\subsection{BERT-Based Model}
As shown in Table \ref{tab:epochs_table}, the recall generally exceeds 75\% across different frame types. Notably, the frame-wise label accuracies achieved by the fine-tuned model outperform the LLM-prompting. In addition to higher accuracy, the model’s predictions exhibited lower standard deviation, further validating the consistency of the multi-task learning approach.

Upon evaluating the fine-tuning loss curve over 30 epochs, we observe that the fine-tuned model's performance gradually improves and stabilizes around the 10th epoch, achieving a balanced performance without compromising accuracy for any specific frame type. An analysis of confusion matrices over epochs reveals particular patterns of misclassifications. For example, early in training, the model confuses financial frames with social ones, a distinction that improves over time. 

\subsection{Large-scale Frame Analysis}
Using our fine-tuned BERT, we obtain frame labels for the entire 760k chunks (19k podcasts). This allows us to examine if the automatic frame labels are consistent with plausible framing associations grounded in real-world patterns. The entity-frame distributions are outlined in Figure \ref{fig:fine_tuned_frame_distribution}. 

Across all 760k chunks (19k podcasts), the \textit{social} frame is the most dominant one ($\approx$60\%). Despite the overall skew, one can still observe variability in frame assignments at the entity level. Here, topics like `COVID-19' present a complex interplay of \textit{social}, \textit{financial}, and \textit{health} factors, reflecting its socio-economic and public health implications. Meanwhile, topics like `Cryptocurrency' are overwhelmingly framed as \textit{financial} ($\approx$65\%), with minor \textit{social} framing, which reflects the prevailing zeitgeist regarding them. Similarly, Constitutional references are dominated by the \textit{legal} and \textit{security} frames ($\approx$30\% each), reflecting their judicial nature. Meanwhile, the entity `Jesus' across 44k chunks exhibits a strong \textit{moral} framing ($\approx$75\%), with a much lower representation of other frames, reflecting typical moral–religious discourse. In a similar vein, the `Muslim' entity as well has a high \textit{moral} framing of $\approx$45\%, but this is contrasted with $\approx$30\% \textit{security} framing, hinting towards the projection safety debates around Muslims. Overall, these patterns closely mirror real-world associations.

\begin{figure*}[!t]
    \centering
    \includegraphics[width=\textwidth]{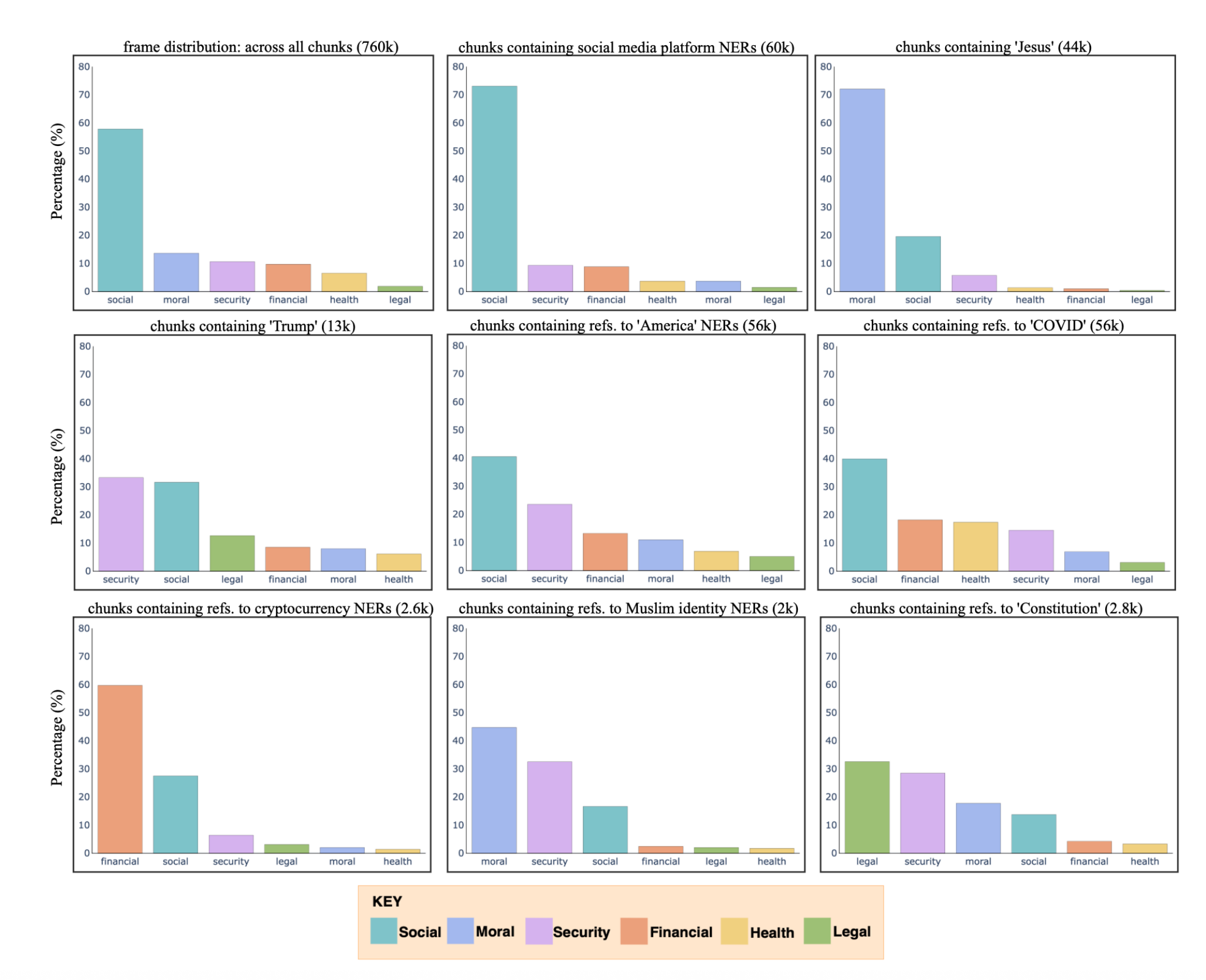}
    \caption{Frame distribution across various NERs using the fine-tuned BERT Model.}
    \label{fig:fine_tuned_frame_distribution}
\end{figure*}

\section{Discussion} \label{sec:discussion}
We conclude our study by revisiting two broad topics: the existing limitations in LLM-prompting for frame analysis and the potential real-world impact of podcasts. 

\subsection{LLM Drawbacks for Automatic Framing} \label{sec:discussion_B}
 Annotators notice interesting patterns that offer insights into the issues LLMs encounter when annotating framing text chunks. A significant concern is context limitation in smaller LLMs. Sometimes, the LLM only extracts key phrases from the start of the chunk, depicting the context limitation. Upon closer examination, we notice that some key phrases are selected only from the beginning 50 words of the chunk.

Secondly, as expected, the LLM outputs are prone to hallucination. For instance, we find that the LLM reported a frame as `Legal' and then hallucinated key phrases like `law,' `regulation, ' `court' (as also seen in Figure \ref{fig:manual_annotation}) to justify its decision, even if these phrases are not present in the chunk at all. This behavior is evident across frame types. While such hallucinations can be checked by exact word match, the annotators also observe a milder form of hallucination occurring due to paraphrasing or misguided importance. For example, in the subsequent text, the key phrase the LLM identifies is `looking forward,' but the chunk only contains `look forward.' \textit{\\ ``...  But slowly but surely we're working our way to be back open as well. So hopefully sooner rather than later. I appreciate that. I don't look forward to trying the end of June, early July to the list. ..." \\} 
Occasionally, even when the frame type is correctly predicted, the LLM directly reports [`phrase1', `phrase2'] as the key phrases, i.e., just as shown in the prompt (see Figure \ref{fig:LLM_frame_prompt}) and demonstrates weak in-context learning ability for subjective tasks. The LLM can also choose specific phrases from the chunk to justify the frame, even if the phrase is insignificant. 
Take the following text for example, \textit{\\ ``... got to fly the T6, that really wasn't accomplishment. I was quite pleased with that. I felt the aircraft was a bit of a time machine. You climb into one of these old war birds and you go fly above the clouds, look at the link tip. There's zero ways to tell that it's 2000, whatever, or 1942. So I thought that was a pretty cool experience. And I tell everybody all the time that out of all the single-engine war birds, there's no question in my mind that T6 is by far the most difficult hurdle. Once you have the T6, I don't want to say mastered because really nobody ever masters the airplane. Once you have it figured out, transitioning to the fighters is easy ..." \\} 
Here, the LLM reports a frame as `Security', focusing on key phrases like `aircraft' and `war birds', while missing the central theme of the chunk -- someone sharing their enjoyable experience of flying a wartime aircraft.

These observations contribute to the research on parametric versus contextual knowledge in LLMs \cite{hagstrom-etal-2025-reality}. \textbf{{The prompted definitions of a frame could not always override the LLM's parametric understanding of specific keywords.}} 

\subsection{Real-World Insights} \label{sec:discussion_A}
As summarised in Table \ref{tab:correlation_analysis_themes_short}, we can draw some interesting topic-word correlations. Within the context of race, we also observe a low correlation between iconic black sports stars (like Michael Jordan and LeBron James) and race in the sports realm, as compared to strong links between race and police and protests. Recent reports also highlight this trend \cite{noauthor_after_nodate,noauthor_nba_nodate}, with implications in downplaying the impact of racial factors in sports. 
A possibly concerning pattern in our analysis is the correlation between racism and comedy within social commentary. Over the last few years, hateful ideas have been amplified, primarily through `concealed punching down' humour \cite{ivry_no_2023,breazu_using_2022}. 

Since \sporc\ includes podcasts released between May and June 2020, the podcast discussions aligned with the worldwide onset of the first-wave COVID-19 pandemic. We observe an apparent rise in conversations about buying medical insurance and investing in healthcare policies. Parallel research has showcased that the pandemic led to a sharp increase in daily purchases of both health and life insurance \cite{chen_pandemic_2023}. Our findings also show a correlation between conversations about loneliness and COVID-19. Other studies have found similar patterns, especially among younger people \cite{rebechi_loneliness_2024}. {\textbf{A striking insight from this analysis is that while loneliness emerged as a long-term consequence of the pandemic, early signs of this were evident through podcast discourse.}}

\section{Conclusion}
This study investigates the effectiveness and limitations of computational modeling in detecting nuanced frames within \sporc. Our experiments suggest a key difference in how LLMs and humans annotate text, where LLMs rely on surface-level features, and human annotators tend to gravitate toward more abstract and interpretive cues. Topic modeling and entity-correlation analyses further reveal thematic overlaps in the discourse, illustrating that podcasts are not only spaces where distinct domains intersect but are also active sites of narrative expansion.

These findings, however, must be considered in light of several constraints. Human annotations for framing remain subjective. Our manual gold set, though representative, is limited to 600 samples. The relatively small train–test size also meant that frame-level efficacy sometimes oscillated over epochs. Scale poses another challenge: working with 1.1M samples under the current compute setup was infeasible. While representative sampling helps ensure that our observations hold at scale, transcribed advertisements introduce additional ambiguities, indicating the need for more robust filtering \cite{abdessamed-etal-2024-identifying} in future versions of \sporc. Looking forward, we aim to incorporate richer relational features in modeling frames and context switching to better understand how narratives evolve.

\bibliographystyle{IEEEtran}
\bibliography{draft_ieeeTCSS}

\end{document}